# Exploring the Law of Numbers: Evidence from China's Real Estate

**Fuqian Zhang; Zhenhua Wang***

School of Information Resource Management, Renmin University of China, Beijing, China
{zfq0906; zhenhua.wang}@ruc.edu.cn

*Abstract*: The renowned proverb "Numbers don't lie" underscores the reliability and insight that lie beneath numbers, a concept of undisputed importance, especially in economics and finance etc. Despite the prosperity of Benford's Law in the first digit analysis, its scope fails to remain comprehensiveness when it comes to deciphering the laws of number. This paper delves into number's laws by taking the financial statements of China's real estate as a representative, quantitatively study not only the first digit, but also depict the other two dimensions of numbers: frequency and length. The research outcomes transcend mere reservations about data manipulation and open the door to discussions surrounding number diversity and the delineation of the usage insights. This study wields both economic significance and the capacity to foster a deeper comprehension of numerical phenomena.

## 1. INTRODUCTION

In the contemporary discourse concerning facets encompassing corporate banking credit, investment due diligence, supplier admission audits, and national regulatory governance, the indispensable role is assumed by the research of their financial statements [1-3]. Financial statements constitute written documents that encapsulate an enterprise's financial position and operational achievements, meticulously chronicling aspects such as assets and liabilities, profitability, cash flows, and changes in owner's equity. These statements offer an illuminating portrayal of the enterprise's developmental and operational landscape, thereby empowering managers to augment their administrative acumen. Simultaneously, they provide investors and creditors with an intuitively accessible means to grasp requisite financial insights, prognosticate prospective corporate profitability, delineate the trajectory of enterprise growth, and underpin wise decision-making on quality of production and operational endeavors [4-6].

The financial statements can be aptly likened to number ecosystems, replete with inherent promise, offering substantive base for the evaluation of their qualitative aspects such as reliability and veracity [7, 8]. This impetus is predicated upon the discernment that ethical constructs and instances of deception transcend the realms of individual or universal paradigms, instead finding their nuanced moorings within distinct sociohistorical milieus [9]. Analogously, the manipulation of the fidelity of numerical data also orbits within this sphere of contextualization [10]. Under this contemplation, Benford's Law has garnered considerable acclaim in existing research, focusing on its congruence with the first digits of numerical sequences [11]. Specifically, the probabilities of occurrence for digits 1 through 9 conform to a logarithmic distribution, with the probability of "1" manifesting at 30.1%, followed by "2" at 17.6%, and progressively diminishing in sequential descent until "9" is encountered, registering a mere 4.6% likelihood.

The Benford's Law finds particular resonance when juxtaposed against financial statements, with its applicability observed across various domains such as Local election [10], hospitality industry [12] and tax audit [13], etc. [11]. However, in the context of China's real estate (CRE), the adherence to this law remains a point yet to be definitively elucidated. The impetus behind this research inquiry resides in a fact: the real estate as the lifeblood of China's economy [14]. To grasp this assertion, one can readily discern it through the nation's expansive populace and entrenched traditional ideologies. According to statistics from the National Bureau of Statistics, the value of real estate has already surpassed 20 trillion, and merely within the initial half of the year 2023, in the immediate aftermath of loosening pandemic containment measures, the investment in real estate development has already surged to a staggering 6 trillion. Consequently, a dedicated examination of number within CRE financial statements not only carries paramount significance for economic advancement but also holds the potential to expand the frontiers of our comprehension of patterns within numerical phenomena. Naturally, this brings the emergence of RQ1.

**RQ1：Does CRE number's first digit follow Benford's Law?**

While Benford's Law offers a window into the distribution of first digits, it is imperative to acknowledge that confining our focus solely to this aspect might entail certain limitations in our understanding. By predominantly fixating on the first digit, we risk inadvertently overlooking a more holistic comprehension of numerical dynamics. Therefore, our scope extends to the numbers themselves, with the emphasis being occupied by the frequency-rank (*f-r*) phenomenon, an eminent distribution that commands chief stage. The significance of delving into the *f-r* emanates from the fact that it encapsulates the fundamental nature of number occurrences. Interestingly, *f-r* is often aptly encapsulated by Zipf's Law [15, 16], a manifestation of human cognitive propensities that resonates universally. Given this, we conjecture that Zipf's Law applies to numbers in a parallel manner. Hence, RQ2 arises self-evident.

**RQ2: Does CRE number's frequency adhere to Zipf's Law?**

Simultaneously, the concept of length can cast light on the nature of objects [17]. From the macrocosmic lifespan of celestial bodies in the natural cosmos, to the microscopic decay length of radioactive elements, and even extends to the growth length of cells in the life, tangible evidence abounds [18-20]. In this vein, the *f-r* law of number length, can reflect the dynamics of CRE — an avenue that, thus far, remains relatively uncharted within previous research endeavors. A natural perception is that numbers with greater lengths are often sparse, as they are mainly reflected in columns such as "total". Conversely, due to the existence of diverse financial details, shorter numbers are more prevalent. Hence, one can envision that the length distribution is likely to exhibit a long-tail phenomenon. Furthermore, due to the unit-price (often with a length of 5) and dimension (often with a length of 3) of CRE, length distribution could potentially feature a left-skewed peak. Whereupon, we posit that the Gamma distribution emerges as a powerful candidate. There is RQ3.

**RQ3: Does real estate number's length comply with Gamma distribution?**

By analyzing the financial reports spanning five years from the top five real estate in China, we answer the above triad of questions, concurrently embarking on a systematic exploration of numerical phenomena. This comprehensive endeavor yields invaluable insights, extending well beyond mere speculations regarding potential data manipulation within Longfor Group's financial records. We also delve into thought-provoking discussions surrounding the maximal diversity in number. Moreover, our delineation of the thresholds dictating number utilization significantly amplifies our comprehension of the intricacies of human cognition. Our contributions are as follows.

1. Three dimensions have expanded the law exploration of number: first digit, frequency, and length.

2. The financial statements of major real estate are considered and extensive experiments are conducted.

3. The radiation and coverage of Benford's law, Zipf's law, and Gamma distribution are extended.

4. The manipulation, maximum diversity, and usage boundary of numbers are discussed.

The remainder of the paper is organized as follows. Section 2 introduces the related work on the three patterns. Section 3 outlines the method. The outcomes and key findings are presented in Section 4. Section 5 is the discussions surrounding the obtained results. Section 7 concludes the paper.

## 2. RELATED WORK

### 2.1. Benford's law

Benford's Law is expressed as Eq.1.

$$f(x) = \log_{10}(1 + \frac{1}{x}) \qquad (1)$$

Here, $x$ represents the first (leading) digit of a number, falling within the set {1, ... ,9}. This signifies that the first digits are not distributed uniformly as one might naively expect, but rather conform to a logarithmic distribution, concentrating more heavily towards smaller values [21]. The frequencies of lower digits (1 and 2) surpass those of higher digits (8 and 9).

Benford's Law goes beyond the purview of mere mathematical inquisitiveness. Its influence has extended into diverse scientific terrains, making the presence felt in renowned distributions like the Boltzmann-Gibbs, Bose-Einstein, and Fermi-Dirac distributions [22]. These manifestations underscore that Benford's Law isn't an isolated equation but rather a foundational trait of physics, occupying a ubiquity niche in unraveling the natural phenomena [23].

Therefore, the law can yield objective insights, exampled by data authenticity and integrity. Departures from the expected distribution, when significant, imply a potential breach in data's veracity — a hallmark of compromised data quality or, more alarmingly, a harbinger of manipulation and deception. This diagnostic capacity positions Benford's Law as a sentinel, entrusted with reminder of irregularities. Consequently, Benford's Law extends far and wide. It has found its analytical imprint in fields as diverse as income recognition [24], to environmental evaluation [25]. It resonates in the throes of local election, illuminating any deviations from expected voter counts, and casts its discerning gaze upon financial transactions, offering an essential lens in the rigorous audit processes of the hospitality industry and taxation domain [10, 12, 13, 26]. Moreover, its observation transcends human constructs, finding resonance in the natural world [11, 23, 27].

Nevertheless, the research of Benford's Law in scrutinizing the financial landscape of China's real estate has thus far remained relatively scarce. This is a noteworthy gap, considering that this industry represents an enormous economic domain, akin to an immense confection. This uncharted terrain presents an avenue for law exploration.

### 2.2. Zipf's law

Zipf's law is succinctly expressed by: the object size $f(r)$ and its rank $r$ adhere to Equ.2 [15].

$$f(r, \alpha, C) = \frac{C}{r^{\alpha}} \qquad (2)$$

For several decades, plentiful domains have engaged in quantitative examinations of Zipf's Law [28-36]. They perceive themselves as complex systems comprised of numerous individual objects whose behaviors interdepend and evolve dynamically, thus giving rise to emergent collective behaviors [37].

Zipf's law is ubiquitous for all systems composed of a large number of living agents, no matter what are their nature and behavior, including human beings, animals or objects recipient of effort of living agents and representing their fulfillment. Whenever a large number of living agents try to achieve something, Zipf law takes place in relation with the output or fulfillment.

Zipf gave a first interpretation of this decreasing rank distribution law by postulating that all human beings minimize effort in their activities to get some fulfillment [16]. This rule was first formulated in 1894 by Ferrero in his paper discussing the mental inertia of human being [38]. But Zipf was the first to explore its possible application to quantitative study. He wrote: "The power laws in linguistics and in other human systems reflect an economical rule: everything carried out by human being and other biological entities must be done with least effort (at least statistically)". This rule is an intuition from the observation of the behaviors of human being himself and probably of other animals, always trying to get more done by doing less.

The concept of minimal effort exerted by human and animal systems is profoundly intriguing, particularly when viewed through the lens of quantitative methodologies. In general, the quest for higher income often necessitates increased effort (time, investment, physical effort etc.) [39]. Yet, whether this principle extends to numbers remains an enigma. This paper aims to unravel this uncertainty, shedding light on whether numbers align with this efficiency-driven phenomenon, or if they deviate from such expectations.

### 2.3. Gamma distribution

If the density function of the random variable $X$ satisfies Eq.3,

$$f(x, \beta, \tau) = \lambda e^{-\beta x} x^{\tau-1}, x > 0 \qquad (3)$$

then $X$ is said to follow a Gamma distribution, denoted as $X \sim Gamma(\beta, \tau)$ [40]. Where, $\Gamma(\tau)$ represents the Gamma function, $\lambda$ is $\beta^{\tau}/\Gamma(\tau)$, $\tau$ and $\beta$ are the shape and scale parameters, respectively. $\tau$ plays a crucial role in determining the shape of the distribution. When $\tau$ is relatively small, the distribution is more heavily skewed to the right. As $\tau$ increases, the distribution becomes more symmetric and approaches a normal distribution when τ is large. β is in determining the spread or concentration of the distribution. It influences the average or expected value of the random variable and affects the shape of the distribution curve. A smaller value of β corresponds to a more concentrated distribution, where the probability mass is concentrated around the mean. In other words, the random variable has less variability and is more likely to take values close to the expected value. As β increases, the distribution becomes more spread out, and the variability of the random variable increases. The Gamma distribution is often used to model positive-valued variables that exhibit right-skewness, where, the skewness is a measure of the asymmetry of a probability distribution, indicating whether the data is concentrated more towards the left or the right side of the distribution. This means that there is a higher probability of observing smaller values, while larger values occur less frequently but can extend to very large values.

A plethora of phenomena in the nature and life exhibit indications of Gamma distribution. Spanning from simulating the lifetimes and decay of cosmic stars and radioactive elements, to the polygonal networks [20] and turbulence [41] in natural systems, to the population growth dynamics within ecosystems, to the infectious diseases [42] and digestion [43] in biological mechanisms, to the development [44], structure [45], and evolution [46, 47] of proteins, and even to the growth, decay, and renewal processes within cells [18, 19].

The frequency-rank faced with number length provide glimpses into the cognitive processes and intellectual deliberations of human beings. It may help track the truth of the digital world and expand scientific understanding.

## 3. METHOD

### 3.1. Data description

The annual financial statements of the five largest real estate enterprises in China have selected, spanning the last five years. These enterprises include Vanke, Evergrande, Country-garden, Longfor, and Sunac. Note that, due to Evergrande's financial crisis in 2022, its financial statements for that year are unavailable. Thus, its data spans from 2017 to 2021, and the other enterprises' data covers the period from 2018 to 2022. In the process of statistical analysis, we opt for integer values, filtering out decimal fractions. Additionally, we conducted a manual review to eliminate anomalous digits, such as those with formats resembling "xxx'" superscripted with a numeral 't'. The results are presented in Table 1. To comprehensively encapsulate financial statements and the ubiquity of numerical occurrences, our selection criteria extended beyond monetary values alone, and various other numbers are embraced. Thus, need to clarify that the number with maximum value (the largest number) doesn't exclusively pertain to monetary figures, as some numbers are expressed in units of "thousands" or "millions".

Table 1: Statistics of numbers.

| Dataset | Observation | Max | Min | Mean | Median |
|---|---|---|---|---|---|
| Longfor2022 | 15458 | 327524894000 | 0 | 283375085 | 100 |
| Longfor2021 | 15530 | 327524894000 | 0 | 290569898 | 100 |
| Longfor2020 | 14223 | 322276158000 | 0 | 291476302 | 100 |
| Longfor2019 | 14236 | 296503846000 | 0 | 144662800 | 127 |
| Longfor2018 | 13468 | 11479336000 | 0 | 53925857 | 102 |
| Evergrande2021 | 3936 | 18000000000 | 0 | 65287889 | 238 |
| Evergrande2020 | 8730 | 53000000000 | 0 | 58580712 | 130 |
| Evergrande2019 | 8207 | 18000000000 | 0 | 50521958 | 167 |
| Evergrande2018 | 7712 | 10162119735 | 0 | 5814463607 | 127 |
| Evergrande2017 | 10121 | 18000000000 | 0 | 2227207724 | 100 |
| Country-garden2022 | 12285 | 27637842220 | 0 | 53316461 | 168 |
| Country-garden2021 | 11799 | 23148390946 | 0 | 52544144 | 228 |
| Country-garden2020 | 13196 | 22035408726 | 0 | 4170638234 | 173 |
| Country-garden2019 | 13574 | 21844661996 | 0 | 19617303 | 179 |
| Country-garden2018 | 10664 | 21740933140 | 0 | 79070199 | 234 |
| Sunac2022 | 9791 | 5448883911 | 0 | 8979315 | 9544 |
| Sunac2021 | 6697 | 4996883911 | 0 | 14798516 | 167 |
| Sunac2020 | 7178 | 4663185911 | 0 | 24825326 | 109 |
| Sunac2019 | 6620 | 4451928611 | 0 | 15724283 | 270 |
| Sunac2018 | 6697 | 4406133709 | 0 | 14411843 | 171 |
| Vanke2022 | 33871 | 91440300192181490 | 0 | 2704541576602 | 92 |
| Vanke2021 | 36262 | 91440300192181490 | 0 | 2526232821114 | 94 |
| Vanke2020 | 35138 | 91440300192181490 | 0 | 2606769225481 | 94 |
| Vanke2019 | 32845 | 91440300192181490 | 0 | 2788451451361 | 95 |
| Vanke2018 | 33353 | 91440300192181490 | 0 | 2745723988603 | 95 |

A phenomenon emerges from the dataset Vanke, involving the most numbers. This could indicate a more extensive scope of business operations and a higher frequency of transactions, which aligns with its position as China's largest real estate developer. Conversely, Evergrande's numbers over the five-year period consistently dwindles, evident from 10121 in 2017 to 3936 in 2021. This conspicuous decline likely mirrors internal challenges, or hint at possible operational turbulence within Evergrande, a viewpoint that is substantiated in 2022, the staggering disclosure of a four-trillion-yuan debt. For other enterprises, maintaining a relatively stable volume, possibly indicative of steady business operations.

It should be explained that Longfor's "*Max*" in 2018 was one order of magnitude lower than in other years, as it did not have a carrying amount held for sale until 2019. Evergrande's "*Max*" in 2020 is much higher than in other years, as this value of 530 million represents issued and fully paid up share capital. The largest numbers involved by Country-garden and Sunac over the past five years are relatively stable, both referring to stock volume. Vanke's 91440300192181490 refers to the unified social credit code.

Moreover, the cardinality of all datasets exceeds 500, substantiating the appropriateness of the data scale. The lowest values in each dataset consistently register at zero. However, the mean values exhibit a degree of unevenness due to the influence of the maximum values. In terms of medians, datasets demonstrate a general trend towards stability, except for Sunac's 2022 median of 9544. This conspicuous divergence suggests a relatively centralized distribution, indirectly hinting that its distribution lacks a relatively pronounced heavy-tail phenomenon.

### 3.2. Regression & Metric

Regression is a ubiquitous statistical way employed to model the intricate relationship between variables by ascertaining the optimal consistency with observed data. It is commonly used to understand how changes in the independent variables affect the dependent variable and to make predictions or forecasts. Its goal is to estimate the parameters of a mathematical equation (the regression model) that best represents the relationship between the variables. To identify the optimal-fit results, the least square method is employed, minimizing the collective discrepancy between the observed values and the predicted values. The determination coefficient, denoted as $R^2$ and confined to the interval of 0 to 1, represents the goodness-of-fit, a pivotal metric. Typically, when an $R^2$ exceeds 0.9, it signifies a strong fit between the regression model and the observed data, suggesting a high degree of conformity. An $R^2$ exceeding 0.8 indicates a reasonably acceptable fit, while values below these thresholds may indicate a lack of compliance with the underlying distribution [35]. It is important to emphasize that $R^2$ alone [48] is often sufficient to assess the level of agreement between the fitted model and the actual data, and has proven effective in numerous quantitative studies [15, 28, 49]. Assuming $p$ and $q$ are observed and fitted values, respectively, and $N$ is the size of the dataset. Then $R^2$ is calculated by Eq.4.

$$R^2 = 1 - \frac{\sum_{i=1}^{N}(p_i - q_i)^2}{\sum_{i=1}^{N}(p_i - avg(p_i))^2} \qquad (4)$$

While $R^2$ is quite informative [48], it should be acknowledged that real-world data seldom adheres rigorously to distributions. Therefore, to bolster the assessment of acceptability, we also supplement our analysis with additional metrics from diverse perspectives.

The Kullback-Leibler divergence (KL) is a widely utilized measure to gauge the similarity between two distributions by quantifying the difference between their predicted and actual distributions [50]. See Eq.5, it ranges from $[0, \infty)$, equating to zero only when the two are identical ($p = q$). A smaller KL value signifies a higher degree of consistency [51]. Typically, a value less than 0.5 is considered acceptable.

$$KL = \sum_{i=1}^{N} p_i \log \frac{p_i}{q_i} \qquad (5)$$

Similarly, the Jensen-Shannon divergence (JS) [52], another extensively utilized measure, and ranges within [0, 1], see Eq.6. A lower value indicates a higher degree of concordance. In practice, a value below 0.2 is often deemed appropriate [53].

$$JS = \frac{1}{2} \sum_{i=1}^{N} p_i \log \frac{p_i}{p_i + q_i} + \frac{1}{2} \sum_{i=1}^{N} q_i \log \frac{q_i}{p_i + q_i} + \log 2 \qquad (6)$$

The final metric, the mean absolute percentage error (MAPE), quantifies the average absolute percentage discrepancy between fitted and actual values see Eq.7. Its range spans [0, +∞), with values below 0.5 generally considered reasonable [54].

$$MAPE = \frac{1}{N} \sum_{i=1}^{N} \left| \frac{q_i - p_i}{p_i} \right| \times (100\%) \qquad (7)$$

In brief, $R^2$, KL, JS, and MAPE are our metrics, where, $R^2$ occupies a primary position due to its direct reflection of the fit quality, and the subsequent three metrics contribute additional viewpoints that enrich the evaluation scope.

## 4. RESULT

This section introduces the answers to the **RQ1 – RQ3**, which are presented separately in Section 4.1 – 4.3.

### 4.1. Number first digit & Benford's law

The visual representation in Fig.1 facilitates the assessment of the alignment between the first digits of the 25 examined datasets (depicted as histograms, from left to right representing digits 1-9) and Benford's Law (illustrated as the blue line). The specific outcomes of the various metrics are reported in Table 2.

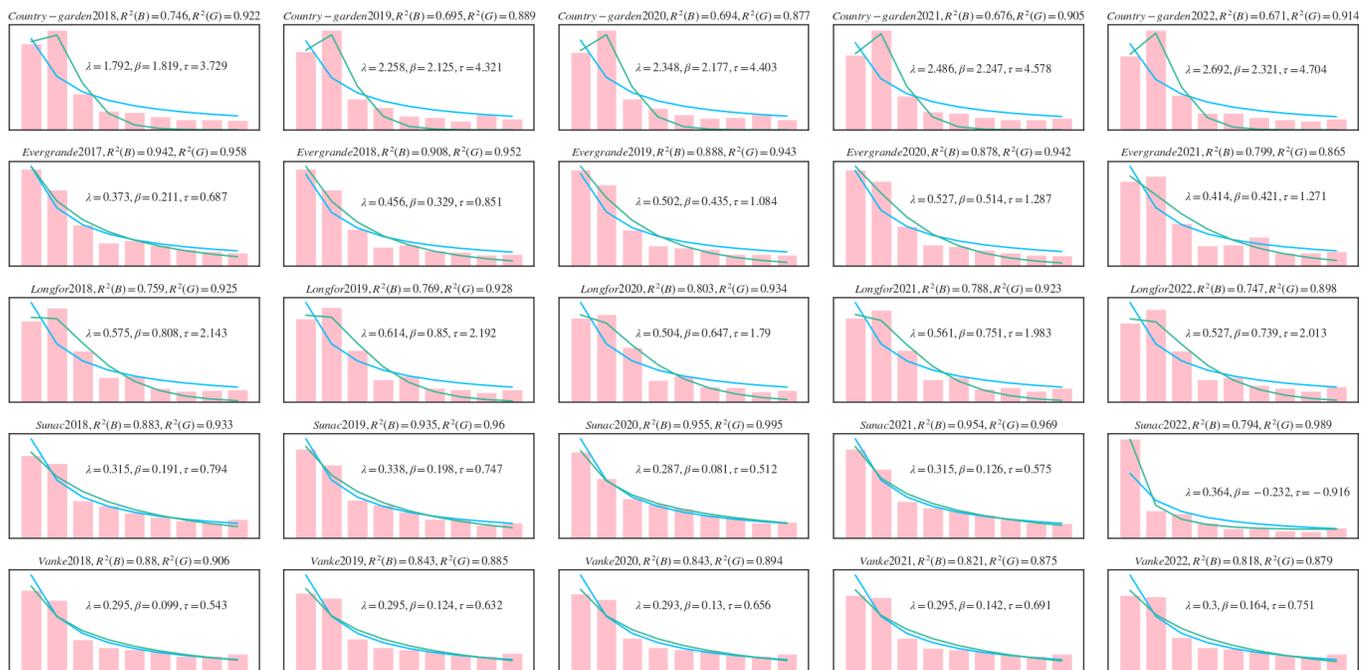

Fig.1: Visualization of first digit distribution. The blue and green lines represent **B**enford's Law and **G**amma distribution, respectively, with the goodness of fit manifested through $R^2(\mathbf{B})$ and $R^2(\mathbf{G})$.

Within the confines of Country-garden, the measure to KL, JS, and MAPE falls relatively high values. Additionally, the $R^2$ value demonstrates a lamentably inadequate alignment. This divergence from anticipated norms is further accentuated through a visual analysis, which distinctly unveils results that veer away from congruity. Notably, the digit "2" raises pronounced suspicions, as its frequency over the five-year span surpasses that of the digit "1," an incongruity that stands in contradiction to Benford's conjecture. Furthermore, conspicuous discrepancies emerge in other aspects. Examples include the occurrences of the digit "8" in both 2019 and 2020, along with the digit "9" in 2021 and 2022. The goodness of fit also corroborates the assertion that Country-garden's dataset diverges from the anticipated Benford's Law. An intriguing phenomenon warrants contemplation: over the five-year trajectory, R2(B) exhibits a consistent descent. This diminishing trend insinuates a growing departure from adherence to Benford's Law. Such observations raise the conjecture that potential data quality issues might be at play. Remarkably, this suspicion finds unexpected validation in the 2023 revelation of Country-garden's staggering debt crisis.

Regarding Evergrande, a conspicuous overabundance of the digit "2" is evident, and furthermore, the digits "4" and "5" deviate from the anticipated pattern in the initial two years. The appearance of the digit "6" in 2021 is also notably aberrant. However, in aggregate, $R^2$, KL, JS, and MAPE collectively suggest that Evergrande broadly adheres to Benford's Law. This paradoxical observation raises intrigue as Benford's Law seemingly becomes less sensitive in discerning data quality for Evergrande, given the its bankruptcy announcement in 2022. A plausible clue lies in the progressively declining $R^2$ over the years, plummeting from 0.942 in 2017 to 0.799 in 2021 (a subpar level). This diminishing goodness of fit hints at potential financial concerns within Evergrande.

Over the span of five years, Longfor's performance parallels that of Country-garden in both visual representation and four evaluation metrics. This divergence from Benford behavior is notably underscored by the digit "2," exhibiting a similar departure from the expected frequency. Additionally, digits "4" and "5" sustain a comparable aura of suspicion. Consequently, a cloud of doubt hovers over the quality of Longfor's financial statements. However, unlike Country-garden, Longfor's fluctuations in the frequency of different digits are less pronounced, hence suggesting that, as of now, the enterprise has yet to manifest overt anomalies.

Despite deviations in the occurrence of digits "1" and "2" for Sunac and Vanke, their overall conformity to Benford's Law is satisfactory, as evident in their $R^2$ values, which consistently exceed 0.8 (excluding Sunac2022). Moreover, their JS values even dip below the thousandth mark, signifying a favorable alignment. In short, they generally conform to the expected patterns outlined by Benford's Law.

Table 2: The fitting results.

| Dataset / Metric | Benford's law | | | | Gamma distribution | | | |
|---|---|---|---|---|---|---|---|---|
| | R2 | KL | JS | MAPE | R2 | KL | JS | MAPE |
| Country-garden2018 | 0.746 | 0.079 | 0.027 | 0.397 | 0.922 | 0.481 | 0.075 | 0.564 |
| Country-garden2019 | 0.695 | 0.084 | 0.029 | 0.364 | 0.889 | 0.740 | 0.095 | 0.616 |
| Country-garden2020 | 0.694 | 0.079 | 0.027 | 0.329 | 0.877 | 0.804 | 0.102 | 0.621 |
| Country-garden2021 | 0.676 | 0.088 | 0.030 | 0.366 | 0.905 | 0.742 | 0.093 | 0.598 |
| Country-garden2022 | 0.671 | 0.095 | 0.032 | 0.397 | 0.914 | 0.742 | 0.091 | 0.593 |
| Evergrande2017 | 0.942 | 0.013 | 0.005 | 0.124 | 0.958 | 0.013 | 0.005 | 0.157 |
| Evergrande2018 | 0.908 | 0.028 | 0.010 | 0.239 | 0.952 | 0.028 | 0.010 | 0.242 |
| Evergrande2019 | 0.888 | 0.034 | 0.012 | 0.263 | 0.943 | 0.041 | 0.014 | 0.298 |
| Evergrande2020 | 0.878 | 0.035 | 0.012 | 0.267 | 0.942 | 0.047 | 0.015 | 0.313 |
| Evergrande2021 | 0.799 | 0.041 | 0.014 | 0.249 | 0.865 | 0.053 | 0.018 | 0.326 |
| Longfor2018 | 0.759 | 0.052 | 0.018 | 0.310 | 0.925 | 0.087 | 0.024 | 0.362 |
| Longfor2019 | 0.769 | 0.055 | 0.020 | 0.339 | 0.928 | 0.097 | 0.026 | 0.391 |
| Longfor2020 | 0.803 | 0.044 | 0.016 | 0.293 | 0.934 | 0.054 | 0.017 | 0.316 |
| Longfor2021 | 0.788 | 0.048 | 0.017 | 0.295 | 0.923 | 0.088 | 0.025 | 0.358 |
| Longfor2022 | 0.747 | 0.047 | 0.017 | 0.262 | 0.898 | 0.094 | 0.027 | 0.372 |
| Sunac2018 | 0.883 | 0.013 | 0.005 | 0.096 | 0.933 | 0.014 | 0.005 | 0.128 |
| Sunac2019 | 0.935 | 0.009 | 0.003 | 0.098 | 0.960 | 0.010 | 0.003 | 0.122 |
| Sunac2020 | 0.955 | 0.005 | 0.002 | 0.081 | 0.995 | 0.001 | 0.001 | 0.046 |
| Sunac2021 | 0.954 | 0.006 | 0.002 | 0.073 | 0.969 | 0.005 | 0.002 | 0.080 |
| Sunac2022 | 0.794 | 0.064 | 0.023 | 0.368 | 0.989 | 0.009 | 0.003 | 0.146 |
| Vanke2018 | 0.880 | 0.016 | 0.006 | 0.141 | 0.906 | 0.017 | 0.006 | 0.154 |
| Vanke2019 | 0.843 | 0.02 | 0.007 | 0.146 | 0.885 | 0.020 | 0.007 | 0.162 |
| Vanke2020 | 0.843 | 0.019 | 0.007 | 0.142 | 0.894 | 0.018 | 0.006 | 0.158 |
| Vanke2021 | 0.821 | 0.021 | 0.008 | 0.150 | 0.875 | 0.022 | 0.008 | 0.173 |
| Vanke2022 | 0.818 | 0.021 | 0.007 | 0.138 | 0.879 | 0.021 | 0.007 | 0.168 |

On the whole, Benford's law enjoys the potential to be suitable for real estate. This implies the presence of data manipulation within their financial statements, adding an intriguing layer that resonates with the discourse surrounding speculative economic trends. However, it's essential to underscore the nuances inherent in such assertions. Here "potential" underscores the non-absoluteness of these findings, acknowledging the potential flaws in data collection and processing as a caveat to the conclusion.

Another conceivable interpretation perspective emerges, that China's real estate is distinctive, and might defy the assumptions laid out by Benford's Law. This engenders a new question: what alternative distribution might govern these datasets? Gleaning from the obvious leftward skewness (exemplified by Sunac2022, where skewness is predominantly anchored around the digit "0"), a suitable hypothesis emerges, with Gamma distribution emerging as a strong contender to model the distribution of first digits.

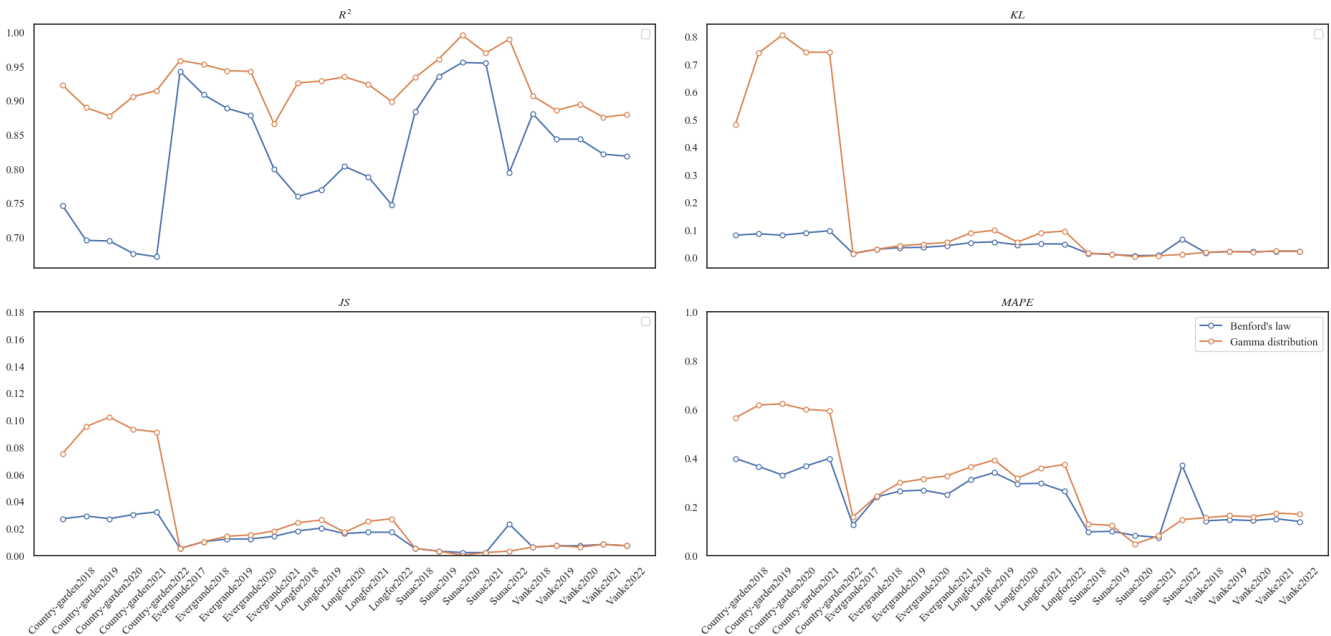

Fig.2: Comparison of Benford's Law and Gamma distribution on four metrics.

The experimental outcomes based on Gamma distribution, as illustrated in Fig.1 and Table 2, serve as a validation of our conjecture, particularly highlighted by the remarkably high goodness of fit values — a pleasantly surprising outcome. Moreover, Fig.2 highlights that, barring the exception of Country-garden (less-than-ideal), the fit results between Gamma distribution and Benford's Law for other datasets are close, signifying a strong alignment. The anomaly or discordance exhibited by Country-garden, conversely, may cause by its data itself, originating from the distortion in its financial concerns. Consequently, within the reviewed time frame, China's real estate appears to adhere more closely to Gamma distribution. Considering the fluctuations in parameters across distinct datasets, a synthesized view over all datasets provides a more indicative perspective, see Eq.8.

$$f(x) = 0.38e^{-0.32x}x^{0.027} \qquad (8)$$

In summation, we can answer RQ1: The first-digit phenomenon can adhere to Benford's Law. Of course, without considering data manipulation, Gamma distribution is more appropriate.

### 4.2. Number frequency & Zipf's law

The visualization presented in Fig.3 enables the assessment of the congruence between the number frequency (scatter points) and the fitting of Zipf's Law (straight line). Fig.4 reports the outcomes of each individual metric.

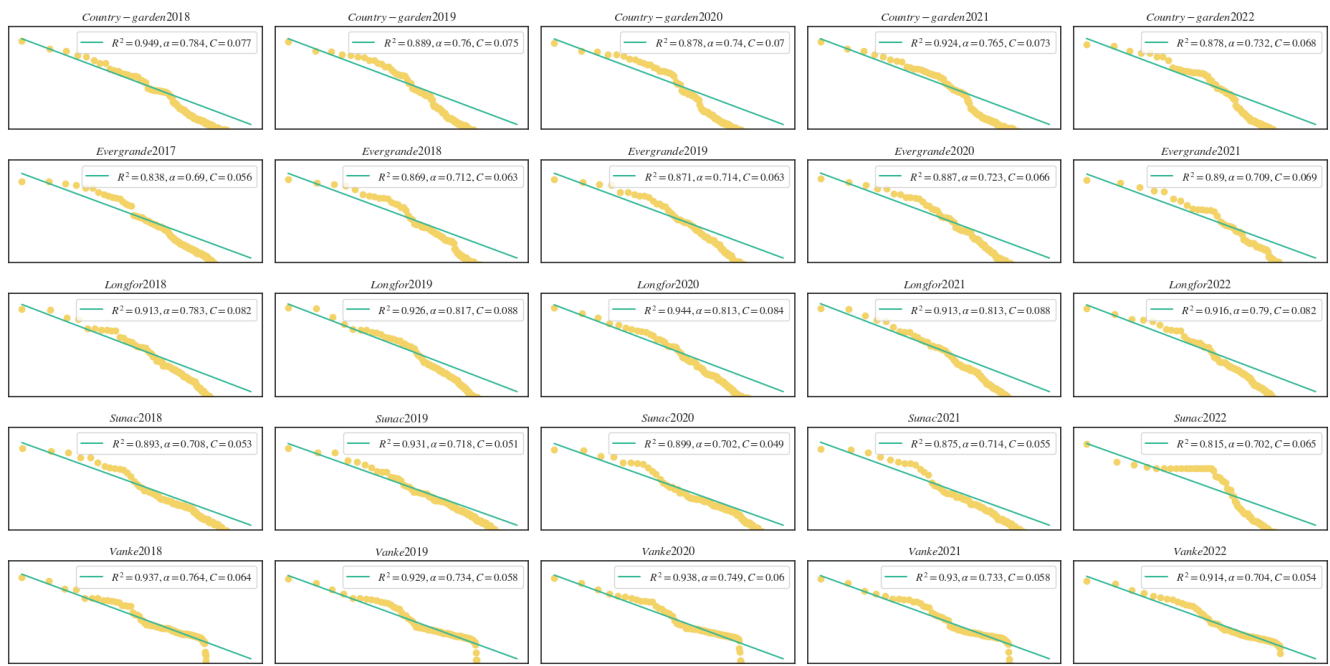

Fig.3: The fit between Zipf's law and number frequency on a logarithmic coordinate system.

The visual representations of consistency across all datasets exhibit encouraging trends, displaying an overarching linear distribution pattern. Notably, all $R^2$ surpass 0.8, with several datasets, including Country-garden2018 and Sunac2019, even achieving $R^2$ exceeding 0.9. These outcomes underscore an alignment of numbers with the of Zipf's declaration. Additionally, while MAPE hovers around an acceptable threshold with some hesitation, denoting a slightly less favorable performance, the KL and JS divergence metrics remain sufficiently small, well below their respective thresholds. This accentuates the acceptability between the actual distribution of numbers and Zipf's Law. Thus, we can claim that number frequency of the real estate is consistent with Zipf's Law.

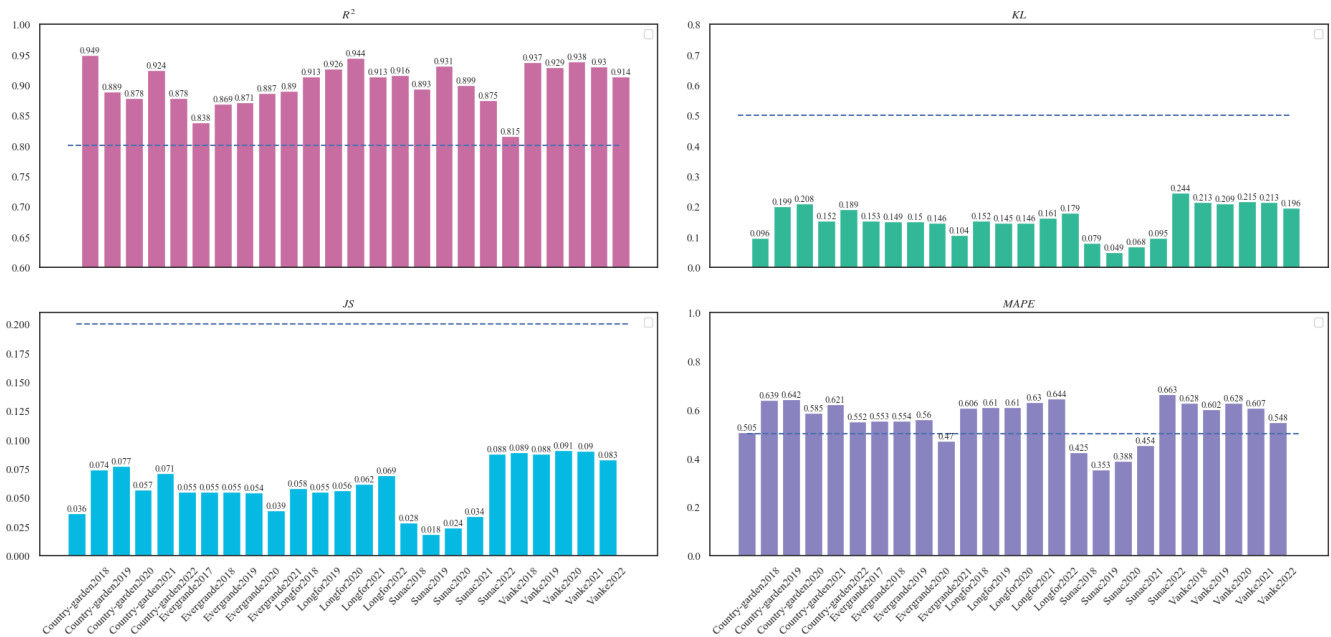

Fig.4: Various metrics for fitting Zipf distribution with number frequency, and the dashed line is the acceptable threshold for each metric.

In our contribution to family of Zipf's Law, we have welcomed numbers as a new member, which can augment our cognitive grasp of the surrounding phenomena. This illustrates the pervasive prevalence of certain numbers juxtaposed with the rarity of others. This paradigm underscores the subtle yet undeniable presence of principle of least effort in our engagement with numbers, regardless of whether we acknowledge it. Our conjecture postulates that this phenomenon can be ascribed, in part, to the strategic endeavor of encapsulating a diverse spectrum of business operations within a streamlined set of numbers, catering to the majority of these transactions.

The observation of parameter variations across distinct datasets prompted us to amalgamate all the datasets, resulting in a specific law, as depicted by Equ.9.

$$f(x) = \frac{0.054}{x^{0.75}} \qquad (9)$$

We discern that the α for numbers rests at 0.75. By juxtaposing this finding with the spectrum of α values characterizing other members of Zipf's family — range from 0.72 for the dialog [35], 1 for the word [15], and 1 for the gene expression [55] to 2 encapsulating the ecological size spectra [56]. Meanwhile, the firm size retains 1 [28], the internet's α gravitates to 0.81 [49], music embodies 0.85[57], citations veer across a range spanning from 2.4 to 3.1 [58], and earthquakes register an α of 1.5 [59].

It becomes evident that the α value for numbers remains relatively discreet. This revelation alludes to a less intensity between the common and the rare that numbers traverse. Furthermore, a profound realization concerning the economic resonance borne by number usage, nurtured by the cognitive reservoir at our disposal. In fact, the α magnitude emerges as a reflection of cognitive resource allocation, invoked to fulfill specific cognitive exigencies [15]. This proffers an intriguing inference — within the context of number engagement, it stands as the least economically driven among its aforementioned counterparts. In tandem, the associated cognitive investment remains proportionally modest. Switch perspective to efficiency, the narrative of number endeavor is marked by parsimony—a propensity to harness effort judiciously for more achieving maximal utility. This prudence may be attributed to the intrinsic information economy underpinning numbers, set in relief against the expanse demanded by other objects like citations and ecological size spectra etc.

### *4.3. Number length & Gamma distribution*

The visualization depicted in Fig.5 allows for appraising the congruence between number length (scatter plots) and the fitted Gamma distribution (red lines). Concurrently, Fig.6 chronicle the outcomes of various assessment metrics.

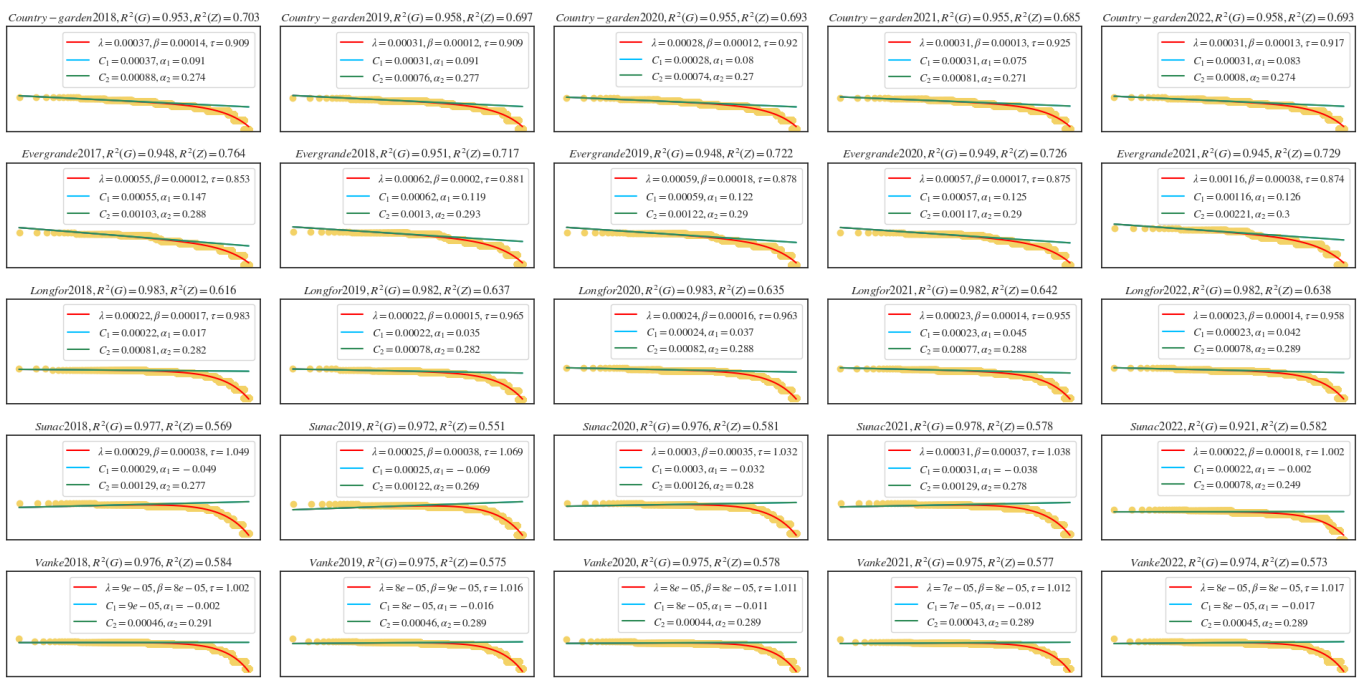

Fig.5: Visualization of number length distribution. The red line represents **G**amma distribution (with parameters λ, β and τ). **Z**ipf's law with different parameters for green and blue lines (each with parameter $C_1$- $\alpha_1$ and $C_2$- $\alpha_2$) The goodness of fit is reflected in $R^2(\mathbf{G})$ and $R^2(\mathbf{Z})$, respectively.

Evidently, the visual representation substantiates the adeptness of Gamma distribution in capturing the patterns of number length. A surprise emerges, where all $R^2$ surpass 0.9, exhibiting a pervasive sense of fitness and accomplishment.

Fig.6 further unfolds the appropriateness within the context of the remaining three assessment indicators, all of which comfortably transcend the predefined acceptability thresholds. This elucidates the compatibility of length with the contours of Gamma distribution. An observation manifests as β tends towards 0, compelling us to preserve precision to five significant figures. This seemingly positions Gamma distribution in a form that approximates the semblance of Zipf's law. In light of this, two experiments are conducted. The first entails evaluating the fit when β was directly set to 0. The second involves an attempt at fitting Zipf's law to the length. They are represented by the blue and green lines in Fig.5. Note that the two visually overlap. However, real data exposes a substantial disparity between its tail behavior and the predicaments posed by the β = 0 scenario. Intriguingly, this inadequacy also befalls the Zipf distribution. Additionally, $R^2$ values conspicuously fail to meet the benchmark. This demonstrates that number length cannot be encapsulated by the supplanting potential of Zipf distribution over Gamma distribution. In conclusion, we can affirm that number length aligns with the contours of Gamma distribution.

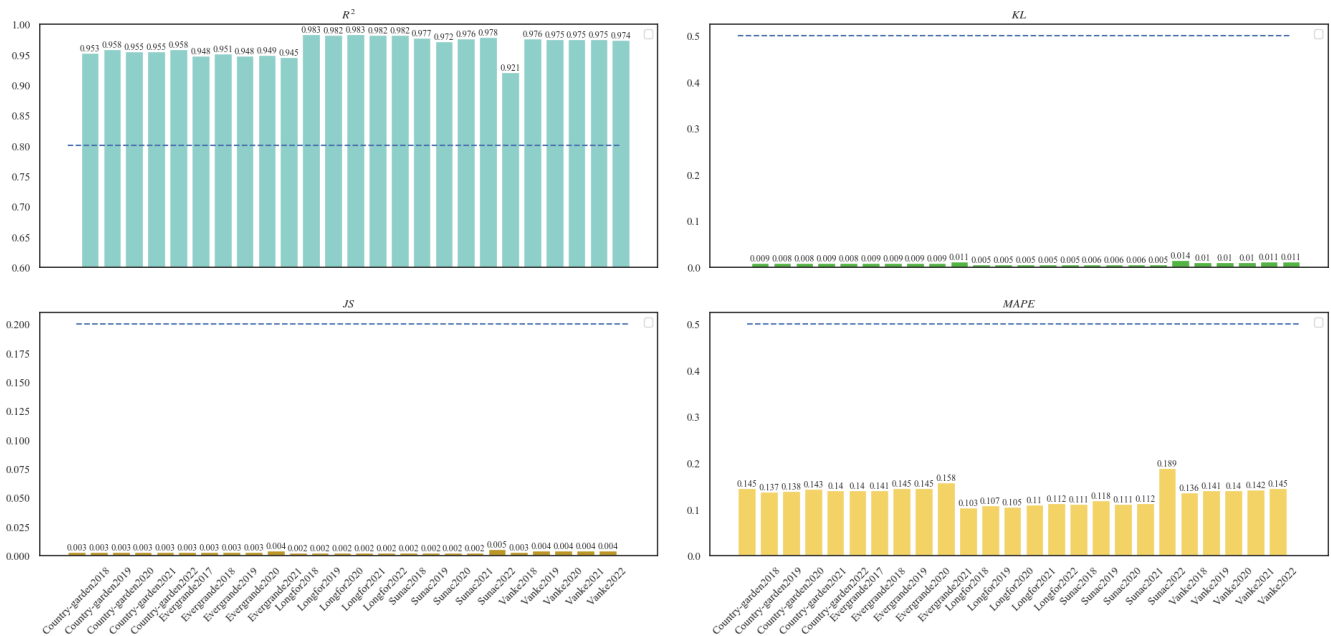

Fig.6: Various metrics for fitting Gamma distribution with number length.

This introspection opens a revelation: the arrangement of lengths showcases a conspicuous long-tail phenomenon. This revelation not only informs us of the prevalence of certain lengths but also underscores the presence of rarer, less frequently employed ones, which appear with a heightened intensity. This contrast alludes to harmonize with the principle of least effort — a minority wields a disproportionately significant impact, that employ a concise set of lengths to meet the majority of requirements. Furthermore, the ubiquity of Gamma distribution across organic systems, underscores its generality. It mirrors a certain vitality and life dynamics. This intriguing parallel lends itself to the notion that numbers, too, might possess a dimension of life. This notion of numbers carrying a temperature, a subtle pulse transcends the conventional perception of them as cold and from icy detachment. Instead, it hints at a result of human deliberation and application on number expressions, where mathematical abstractions are woven into the fabric of human understanding and decision-making.

We coordinate all lengths to characterize specific mathematical expressions, see Eq.10.

$$f(x) = 0.002 e^{-e^{-5}x} x^{-0.049} \qquad (10)$$

We now understand that Gamma distribution is evident in both the first digit of number and its length. In fact, the frequency distribution of the first digit can be regarded as equivalent to its frequency-rank distribution. Alternatively, we can understand it this way: the value of first digit itself can be abstracted as the rank. Therefore, first digit and length are in the same distribution system, with power exponents of 0.027 and -0.049, respectively, highlighting disparate trends between them. The former manifests as a more pronounced polarity between a select few high-frequency lengths and a multitude of less frequently observed ones, which leans towards accentuating a broader spectrum of length diversity, encapsulating the variability in length expressing number. The latter indicates that the occurrences between common first digits and rare ones are relatively less severe and intense, and their concentration and scarcity phenomena are not as prominent as the length.

The trajectory delineated by the distribution curve, underscores the switch of number length, which exhibits hastier at elevated frequencies than first-digit, engendering a tighter aggregation of lengths at the higher frequency spectrum, see Fig.7. Notably, the rapid declination in frequency density transpires expeditiously over a limited set of ranks, progressing from the relatively commonplace to the relatively scarce. This phenomenon implicitly signals a pragmatic endeavor towards heightened efficiency, strategically employing a limited subset of ranks to adequately accommodate the lion's share of requisites. In contrast, the progression from prevalent to scarce instances in first digits unfurls with a smoother trajectory. Within the lower frequency spectrum, both number length and first digit betray a relatively uniform behavior. Despite this, the switch in the former's distribution retains a marginally more intensity than that in the latter. This leads to the inference that the variance in the utilization of first digit is relatively less conspicuous in comparison to length.

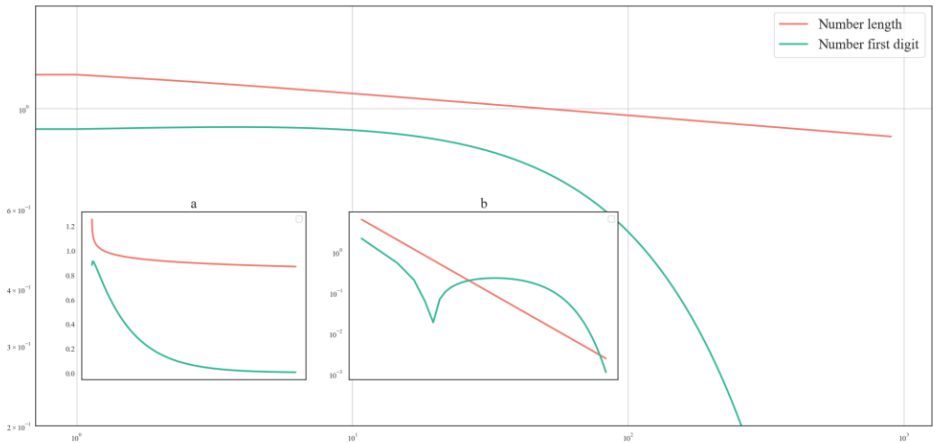

Fig.7: Number length *vs.* first digit. Where, λ is generally regarded as a free parameter, and thus normalized. The overarching graph is presented in a logarithmic coordinate system, while subgraphs "a" and "b" are constructed in standard ones. The latter characterizes the 1-order derivative absolute values of length and first digit.

## 5. DISCCUSSION

We have completed the answer to **RQ1-3** and provided specific law expressions, offering an intuitive understanding.

Then, the following discussions arise.

### 5.1. Upper boundary estimation

De et al., argue that dynamical systems reflected in the distribution of objects often exhibit temporary under-sampling (limited statistical data is inevitable) [60], leading to instability, which manifested as deviations in object scale, allows us to estimate potential upper-cutoff $o_M$ in the system. Building upon this insight, we have an opportunity to explore the boundaries of number.

For Gamma system, λ is generally regarded as a free parameter, and the empirical deviation $D_e$ satisfies:

$$D_e = N(o_m / o_M)^{1/1-\alpha} e^{\beta(o_m - o_M)}$$

Where, $N$ is the number of objects, $o_m$ is the lower-cutoff. The initial value used for iterating $o_M$ is given by the largest observed object in the system.

Then follow:

$$o_M = o_m(1 + D_e / N)^{1/1-\alpha}$$

Since $\alpha \neq 0$, allows $D_e \neq 1$, thereby implying that the $o_M$ does not exhibit divergence. By iterating the aforementioned equations, we can estimate it. Likewise, for Zipf's system:

$$o_M = o_m(N / [N(o_M / o_m)^\alpha - 1])^\alpha$$

There are the following inferential insights

In the absence of any presupposition regarding data manipulation within the China's real estate economic context, we are confronted with an intriguing phenomenon. Our estimations reveal an observation: the maximum frequency at which the first digit appears is 46.15%, exceeding the theoretical upper limit of 31.10% as dictated by Benford's Law. This 46.15% reflects a certain threshold of human tolerance in deploying digits. We could believe that this threshold might be intertwined with the aesthetics of digits. Indeed, the significance of digits 1-9 in terms of symbols is equivalent (which is also why people tend to intuitively assume that first digit meets a uniform distribution before investigation). If first digit occurs too frequently, it might be considered dissonant, and this dissonance boundary might be around 0.4615. Moreover, the first digit serves as the utmost significant digit, indicating the highest order of magnitude within a number. An excessive prevalence of the digit '1' can consequently constrain the spectrum of numerical value expression. This raises pertinent questions about the potential erosion of numerical diversity.

Our estimations further unveil that the frequency of the most frequently used numbers does not surpass 4.06%. Although we know that human cognitive limitations lead to a small amount of numbers being used extensively (For ease of understanding, our law can be simplified through the Pareto like Principle, where the first 20% shoulder the lion's share of utility) —the equilibrium within this select group is not governed by random choice. Instead, it's rigorously maintained to ensure that no single number bears excessive communicative burden. This observation not only underscores the influence of diversity of numbers but also highlights the drive toward efficiency in numerical representation, where overreliance on a single number is judiciously averted.

We estimate that within a set of economic number collections, the occurrence of the lengthiest numbers—those that occupy the apex of numerical magnitude—does not exceed a minute fraction, a mere 0.015%. This pattern evokes the concept of attachment found in ecological systems. Much like the stability observed in an ecological pyramid, where higher organisms are perched at the pyramid's peak, our numerical dataset appears to exhibit a similar stability. Here, the longest numbers occupy the upper echelons of this metaphorical pyramid, also suggesting that numbers may follow a hierarchical structure. Furthermore, this phenomenon resonates with social dynamics. A small percentage of individuals often achieve the highest levels of recognition or power, effectively standing at the peak of the metaphorical pyramid. It's worth considering that this elite fraction could well represent the top 0.015% of individuals from insights of numbers. The occurrence in turn also means that numbers with short lengths are common, which may be attributed to: Firstly, these relatively shorter lengths align with practical usage requirements. In real estate, numbers typically serve to represent quantities such as area, price, and monetary values. These quantities, in most cases, are concise numbers from the beginning. Secondly, the preference for shorter representations is a human tendency rooted in ease of application and memorability. Individuals involved in real estate transactions, contract negotiations, and data management processes often lean towards shorter numbers to streamline their workflows, a phenomenon astutely exemplified by the prevalent practice of quantifying numbers in units such as "ten thousand" and "one million". Also, the limited computational and data storage resources available play practical considerations. Given the digital nature of modern data handling, storage, and transmission, excessively long numbers may impose resource burdens.

### 5.2. Implication

A plethora of studies has underscored the utility of examining the adherence of economic numerical data to Benford's Law as a foundational metric for ascertaining potential artificial data manipulation. Following a rigorous analysis of China's real estate data, we have ascertained that Evergrande, Country-garden, and Longfor do not conform to Benford's Law, thereby signifying plausible financial data manipulation. It is imperative to highlight that this conjecture gains further credibility when we consider the documented financial crises experienced by Evergrande and Country-garden. Considering the situation, there is a significant increase in suspicion of financial manipulation related to the abnormal situation in Longfor. As a prudent course of action, we recommend that relevant regulatory authorities maintain a vigilant oversight of this situation.

Nevertheless, adopting an alternative perspective potentially allows us to uncover less dire implications. Notably, we've identified that, the first digit adheres to Gamma distribution. This revelation unveils a broader spectrum of governing principles, extending the applicability of number laws and enriching our knowledge framework. It also suggests a plausible nexus between the first digit and the "minimum effort principle" wherein smaller digits (such as 1 and 2) exhibit higher frequencies, while larger ones (such as 8 and 9) are relatively less common. It is intriguing to note, however, that the disparities in prevalence between these digits are not particularly pronounced, adding another layer of intrigue to our findings.

While Benford's law serves as a pivotal perspective, its isolated application can unintentionally obscure deeper understanding. Our exploration has unveiled the presence of f-r dynamics within number systems, thus compelling us to broaden the foundation bolstering the significance of numbers and their contribution to the broader realm of Zipf's law.

This revelation not only enriches our comprehension of Zipf's law itself but also extends our understanding of the world around us.

As expounded by Mazzarisi [61] in "Maximal Diversity and Zipf's Law", Zipf's law invariably coincides with the maximization of diversity within component sizes. The underlying principle behind this maximization of diversity might indeed align with the more general concept of the principle of least effort and the least action principle in physics. We believe that in the case of Zipf's law, the diversity of numbers will be elevated to a driving force. This, in turn, results in empirical distributions shared by numbers and numerous other components within natural, societal, and human systems, conceivably bearing strong correlations with the optimization processes observed in the natural evolution of ecosystems. From the perspective of biological mechanisms, it seems to be a tracing back to the origin [62]. Living organisms, *vs.* number self-containing organizations, akin to the orchestrated coordination of multiple heat engines *vs.* multiple digits, which inherently invokes the principle of least action derived from physics, expressible as Hamilton's principle. At their core, these principles converge into mathematical variational principles that delve into extremum problems associated with functionality.

The lengths of numbers are not disorderly but rather adhere to Gamma distribution. While there exist, nuanced mathematical distinctions compared to Zipf's law, the overarching principle of minimizing effort remains a shared essence. We believe that this can be driven in the same direction as Zipf's law, with maximum diversity. These two minimum effort principle systems not only embody resilience and stability but also, as stated by scholar Scott E. Page, yield what he terms the "diversity dividend." Page has substantiated this concept: within systems grappling with challenges demanding innovation and exploration, diversity confers predictive capabilities beyond what the original system's average could attain. We can also understand from that when conceiving number as a communication power system, the pursuit of optimizing overall efficiency, including future endeavors, necessitates not only internal optimization within the sphere of the sender's economy (the point of origination) and the establishment of a unified mechanism but also the strategic utilization of diversification within the receiver's economy (the point of reception and expression). This strategic utilization leads to the emergence of new systems themselves and paths prospective avenues in the future, all achieved with minimal resource consumption costs.

It is well-established, individual information theory that life itself comprises a multitude of measures, enabling the dissemination of its intrinsic information from the past to the future while preserving temporal coherence. Throughout this journey of number, the heterogeneity of digit, frequency and length serves as the wellspring of happiness. In pragmatic terms, the crux of the potency residing in diversity lies in its capacity to engender a multiplicity of perspectives, not merely nominal labels, thereby cultivating "dividends" through super-additivity that transcend the capabilities of any solitary component. This constitutes a foundational, bottom-up process of paramount significance when addressing issues devoid of historical trajectories for reference [63].

This study elucidates that numbers adhere to a more intricate course than what meets the eye. As we navigate the diverse and captivating numerical landscape in the future, we may find ourselves echoing the sentiments of poets, realizing that numbers, too, articulate themselves in the immaculate and efficient manner.

Indeed, this paper is poised to catalyze advancements in engineering technology. One intuitive application lies in financial fraud detection and data manipulation verification [1, 3, 4, 6-8]. It's conceivable that we can boldly embark on collaborative efforts leveraging the insights derived from Benford's law, Zipf's law, and Gamma distribution, all residing within the numerical data. Furthermore, these laws have the potential to invigorate various facets of financial analysis through deep learning. This encompasses the refinement and enhancement of sampling, extraction, classification, recognition, parsing, matching, inference, and an array of other analytical processes [2].

## 6. CONCLUSION

In this paper, empirical evidence from real estate financial statements in China reveals the distribution patterns behind the numbers, covering Benford' law represented by first digit, Zipf's law satisfied by number frequency, and Gamma distribution with consistent number length. Note that without considering data manipulation, the first digit is more in line with Gamma distribution. Furthermore, our research illuminates the "principle of minimum effort" emerges in numbers, which is intricately linked to efficiency and maximal diversity of human use. The resulting revelations within the number usage landscape suggest that numbers possess a certain "temperature" a concept transcending the frigid confines of mere numerical values.

The limitations of this study should be discussed. The real estate in China occupies a rather distinctive niche, and, as such, the generalizability of certain findings may warrant prudence or is slightly insufficient. It is our aspiration that future research endeavors cast a wider net. Note that it remains indisputable that the empirical data selected for this study span a consequential temporal spectrum and encompass an array of data types, thereby conferring a veneer of substantiality and resilience to our findings. On the other hand, despite the first digit, frequency and length have taken important steps in fully understanding and characterizing the laws of numbers. Yet, let us not be myopic: there may exist uncharted dimensions that have eluded our scrutiny, holding the potential to enrich our comprehension of numerical intricacies.